\documentclass[10pt,twocolumn,letterpaper]{article}

\usepackage{wacv}              

\usepackage{graphicx}
\usepackage{amsmath}
\usepackage{amssymb}
\usepackage{booktabs}

\usepackage{bmpsize}

\usepackage[utf8]{inputenc} 
\usepackage[T1]{fontenc}    
\usepackage{url}            
\usepackage{amsfonts}       
\usepackage{nicefrac}       
\usepackage{microtype}      
\usepackage{xcolor}         
\usepackage{amsmath}
\usepackage{amssymb}
\usepackage{multirow}
\usepackage{wrapfig}
\usepackage{pifont} 
\newcommand{\cmark}{\ding{51}} 
\newcommand{\xmark}{\ding{55}} 
\usepackage{caption} 

\usepackage[pagebackref,breaklinks,colorlinks]{hyperref}

\usepackage[capitalize]{cleveref}
\crefname{section}{Sec.}{Secs.}
\Crefname{section}{Section}{Sections}
\Crefname{table}{Table}{Tables}
\crefname{table}{Tab.}{Tabs.}

\def\wacvPaperID{2316}
\def\confName{WACV}
\def\confYear{2025}

\begin{document}

\title{Ada-VE: Training-Free Consistent Video Editing Using Adaptive Motion Prior}

\author{Tanvir Mahmud, Mustafa Munir, Radu Marculescu, and Diana Marculescu\\
The University of Texas at Austin
}
\maketitle

\begin{abstract}
Video-to-video synthesis poses significant challenges in maintaining character consistency, smooth temporal transitions, and preserving visual quality during fast motion. While recent fully cross-frame self-attention mechanisms have improved character consistency across multiple frames, they come with high computational costs and often include redundant operations, especially for videos with higher frame rates. To address these inefficiencies, we propose an adaptive motion-guided cross-frame attention mechanism that selectively reduces redundant computations. This enables a greater number of cross-frame attentions over more frames within the same computational budget, thereby enhancing both video quality and temporal coherence. Our method leverages optical flow to focus on moving regions while sparsely attending to stationary areas, allowing for the joint editing of more frames without increasing computational demands. Traditional frame interpolation techniques struggle with motion blur and flickering in intermediate frames, which compromises visual fidelity. To mitigate this, we introduce KV-caching for jointly edited frames, reusing keys and values across intermediate frames to preserve visual quality and maintain temporal consistency throughout the video. With our adaptive cross-frame self-attention approach, we achieve a threefold increase in the number of keyframes processed compared to existing methods, all within the same computational budget as fully cross-frame attention baselines. This results in significant improvements in prediction accuracy and temporal consistency, outperforming state-of-the-art approaches. Code will be made publicly available at \href{https://github.com/tanvir-utexas/AdaVE/tree/main}{https://github.com/tanvir-utexas/AdaVE/tree/main}.
\end{abstract}

\section{Introduction}
The field of video generation has gained significant attention due to its numerous practical applications, including video editing~\cite{lee2023textvideoedit}, video synthesis~\cite{make_a_video}, and style transfer~\cite{vid_style_transfer}. This paper focuses on video-to-video synthesis, which involves altering videos through text prompts while preserving their original motion and structure~\cite{text2video-zero, geyer2023tokenflow, zhang2023controlvideo}. Recently, video synthesis techniques based on diffusion models have garnered considerable interest for their outstanding performance in image editing and generation tasks~\cite{pnpDiffusion2023, blattmann2023videoldm}. However, training these models from scratch is computationally intensive, data-demanding, and necessitates frequent updates with the release of new models~\cite{liang2023flowvid}. To address these challenges, we explore training-free extension of text-to-image (T2I) diffusion models for video-to-video synthesis tasks, avoiding the need for extensive retraining and leveraging the capabilities of pre-trained models. 

\begin{figure*}[t]
\centering
\includegraphics[width= 0.9\textwidth]{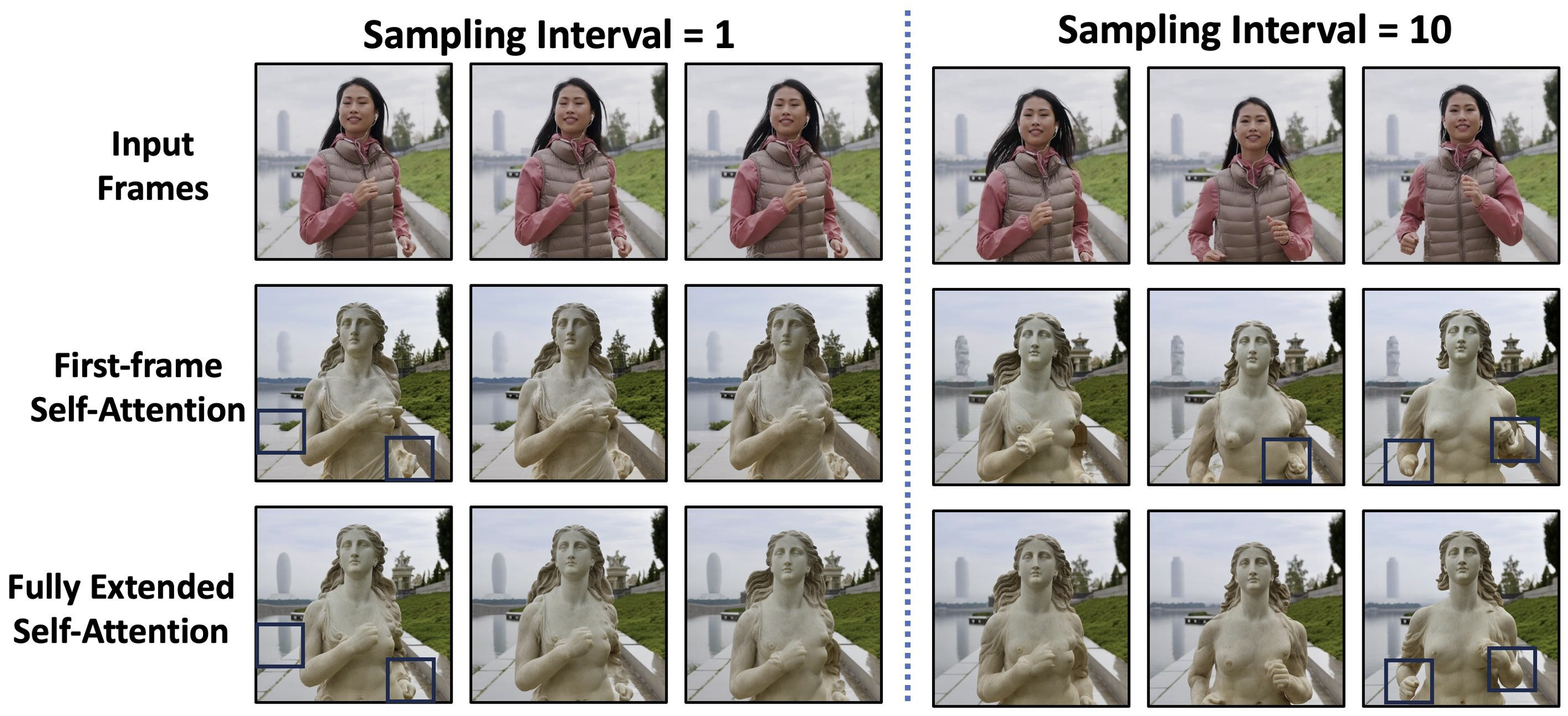}
\caption{
\textbf{Effect of self-attention extension on diverse motion:} Frames are sampled at intervals of 1 (slow motion) and 10 (fast motion). Two methods are compared: one uses the first frame's key-value (KV) across all frames for efficiency, while the other fully extends KVs, which is computationally intensive. The efficient method works well in slow motion but struggles with faster motion, where full extension achieves better results. This highlights the need for adaptive self-attention based on motion to enhance video quality, reduce redundant computations, and incorporate more frames in the self-attention process.}
\label{fig:1}
\end{figure*}

In contrast to image editing, video editing faces several key challenges that hinder its widespread adoption for real-world applications. First, consistent character generation across frames based on text prompts, while preserving the structural details of the guidance video, is crucial. Since diffusion models generate outputs stochastically from noise, achieving consistent character generation over the video is challenging, often resulting in flickering for non-persistent content across successive frames. Second, maintaining temporal coherence in the generated frames is essential for realism, as human eyes are highly sensitive to temporal inconsistencies. Most existing video diffusion methods suffer from blurring effects~\cite{geyer2023tokenflow} and temporal inconsistencies~\cite{zhang2023controlvideo, text2video-zero}, compromising the quality of the video. Lastly, real-world videos exhibit diverse dynamics, including fast motion transitions, necessitating adaptive model operations to maintain quality. A single approach may not be suitable for all video classes in real-world applications, highlighting the need for adaptable and robust methods across different scenarios.

Inflating text-to-image (T2I) synthesis models is a widely used approach to leverage image knowledge for consistent video generation. Several methods have focused on fine-tuning these inflated models to improve performance~\cite{wu2022tuneavideo}. Recently, research has emphasized the training-free adaptation of T2I models for video editing tasks~\cite{qi2023fatezero, text2video-zero, geyer2023tokenflow}. These studies highlight that cross-frame self-attention plays a crucial role in maintaining consistent character generation across frames. Typically, cross-frame self-attention extends inter-frame key-value (KV) pairs for queries from each frame~\cite{geyer2023tokenflow, zhang2023controlvideo}, but this increases the operational complexity quadratically. Common optimization methods include using KV from only the first frame~\cite{text2video-zero}, combining KV from the first and previous frames~\cite{wu2022tuneavideo}, and sparsely sampling KV from the temporal window~\cite{peng2024conditionvideo}. However, the optimal choice of KV depends significantly on the underlying motion dynamics of the video. Empirical observations (see Fig.~\ref{fig:1}) show that slow-moving regions benefit from sparser KV sampling to reuse similar features, while fast-moving regions require denser sampling to capture detailed motion. Heuristic designs of KV extension may be suitable for specific use cases but often lead to redundant computations in slow-moving videos or inconsistent results in fast-moving videos. Thus, balancing self-attention extension is a major challenge for achieving consistent video generation while maintaining quality.

In addition to consistent editing across frames, the video editing objective has additional challenges of achieving temporal/motion consistency in all intermediate frames. Most recent works have focused on two-step approaches for maintaining temporal consistency across the video: first, jointly editing a set of randomly/uniformly sampled key frames over the video, followed by custom intermediate frame-interpolation techniques. Several interpolation methods have been explored, including flow-warping~\cite{liang2023flowvid}, token flow~\cite{geyer2023tokenflow}, and traditional video interpolation techniques~\cite{yang2023rerender}. However, these methods often lead to either blurring or flickering effects, especially in videos with fast motion, which reduces the overall video quality. Another approach involves extracting relevant features from nearby key frames only during the joint editing phase and selectively reusing these features for intermediate frames~\cite{zhang2023controlvideo}. Despite this, such methods typically suffer from inconsistent character generation and temporal inconsistencies, heavily dependent on the feature selection method used. Therefore, the trade-off between motion blurring and temporal inconsistency necessitates more sophisticated methods for robust video editing. Generally, extensive use of cross-frame self-attention can improve performance, but at the cost of increased computational demands, while consistent frame interpolation remains a significant bottleneck for achieving high-quality videos.

In this paper, we propose Adaptive Video-Editing (Ada-VE), a novel approach to address the major challenges in video-to-video synthesis. Ada-VE adaptively integrates essential details in joint cross-frame attention while suppressing redundant parts, thereby increasing the number of frames that can be edited without compromising visual quality or increasing computational burden. To achieve this, we selectively incorporate the most salient key-value pairs (KVs) from the moving regions across frames, while leveraging optical flow estimation to identify these regions. This dynamic extension of KVs in cross-frame attention optimizes self-attention based on the motion dynamics of the video, thus reducing redundant computations without losing details. Additionally, we find that preserving the same KVs across frames is crucial for maintaining inter-frame consistency. Based on this observation, we introduce the \textit{KV-cache technique}, which reuses KVs from cross-frame self-attention in intermediate frames. Both of these methods contributes improving both visual quality and temporal consistency.

Our key contributions can be summarized as follows:
\begin{enumerate}
    \item We propose an adaptive extension of cross-frame self-attention, enabling processing an increased number of jointly-edited frames for better consistency without increasing the computational complexity.
    \item We introduce the use of KV-cache from jointly edited frames to maintain temporal consistency and visual quality in intermediate frames of the video.
    \item We conduct extensive experiments demonstrating the effectiveness of Ada-VE compared to state-of-the-art methods. Notably, with the same computational budget, Ada-VE processes three times more frames in extended cross-frame self-attention, leading to a 45\% reduction in warp error and a 25\% improvement in CLIP-Score.
\end{enumerate}

The remainder of the paper is organized as follows. Section 2 describes the related work in video synthesis and frame interpolation, while Section 3 describes preliminary information on video diffusion models. We describe our methodology in Section 4, and our quantitative and qualitative results in Section 5. Finally, Section 6 discusses the broader impacts of our work, and Section 7 summarizes our findings and limitations.
\section{Related Works}
\subsection{Video Synthesis Methods}
Various GAN-based methods have been explored for video synthesis~\cite{wang2020imaginator,bar2022text2live, gupta2022rv, munoz2021temporal}, but these approaches often encounter significant challenges. They are difficult to train, prone to mode collapse, and typically produce lower-quality results~\cite{beatgan, hong2022improving}. Recently, diffusion probabilistic models (DPMs) have achieved remarkable advancements in image generation and editing~\cite{pnpDiffusion2023, controlnet, bar2023multidiffusion, chefer2023attend, ma2023directed, cao2023masactrl}. Several methods have tried to leverage these image priors for video generation and synthesis~\cite{Ceylan2023Pix2VideoVE, yang2023rerender, gen1, wu2022tuneavideo, Liu2023VideoP2PVE, video_epitome, ho2022imagen_video}. However, video synthesis presents additional challenges compared to image synthesis, particularly in maintaining consistent character generation over temporal windows, which demands substantial computational resources and large-scale training~\cite{gen1, liang2023flowvid}. 

Many approaches have attempted to inflate cross-frame self-attention to utilize image priors for video synthesis~\cite{wu2022tuneavideo, geyer2023tokenflow, zhang2023controlvideo}. However, finding the optimal balance between the number of frames and computational efficiency remains a significant challenge in existing methods.
More recently, SORA has demonstrated surprising results in video synthesis, but it is not available as an open-source tool and requires large-scale training on curated datasets~\cite{liu2024sora}. Despite these advancements, open-source methods often suffer from flickering and blurring effects due to inconsistent character generation, and exhibit reduced visual quality. Our goal is to address these limitations and enhance the performance of video synthesis methods.

\subsection{Temporal Consistency in Video Editing}
Maintaining temporal consistency and preserving the structure of the original video in the generated frames is a primary challenge in existing methods~\cite{lee2023textvideoedit, Khachatryan2023Text2VideoZeroTD, zhang2023controlvideo}. For consistent character generation across multiple frames, diverse heuristic self-attention extension method is introduced~\cite{liang2023flowvid, geyer2023tokenflow, text2video-zero}. Several approaches have introduced flow-warping techniques to generate intermediate frames from jointly edited frames~\cite{yang2023rerender, liang2023flowvid, text2video-zero}. However, flow-warping often introduces errors due to occlusions. Some methods have attempted to train diffusion models as inpainting models to address these limitations, but consistent inpainting remains challenging and requires large-scale training, often resulting in lower visual quality and flickering~\cite{liang2023flowvid}. Instead of using flow-warping for interpolation as prior work, we primarily use optical-flow~\cite{teed2020raft, hui2018liteflownet, ilg2017flownet} based motion prior to coarsely estimate the moving parts to enhance performance, without being constrained by precise optical flow estimation.




Tune-a-Video~\cite{wu2022tuneavideo} proposed fine-tuning the model on a specific video, but this approach sacrifices generalizability. Recently, TokenFlow~\cite{geyer2023tokenflow} introduced a method of flowing self-attention features from key frames based on spatial similarity. Despite its promising performance, TokenFlow often blurs intermediate frames and reduces overall quality. ControlVideo~\cite{zhang2023controlvideo} proposed reusing attention features from nearby frames to improve consistency. However, this method can still result in significant flickering and inconsistent character generation for feature disparity in subsequent frame synthesis.
More recently, Motion-Transfer~\cite{motion_transfer} introduced use of pre-trained text-to-video models for editing, requiring additional training on video data. Moreover, this approach significantly increases computational overhead compared to T2I-based methods, restricting its use to very short video duration. Several real-time video editing methods~\cite{streamv2v, streamdiff} have been introduced, focusing primarily on faster processing with minimal self-attention extension based on predefined heuristics, but these often suffer from flickering and temporal inconsistencies.
Our method focuses on enhancing both consistency and quality by optimizing the adaptive-extension and reuse of self-attention features from joint-editing, aiming to overcome the limitations of previous approaches.

\section{Preliminaries}
\begin{figure*}[t]
\centering
\includegraphics[width= 0.9\textwidth]{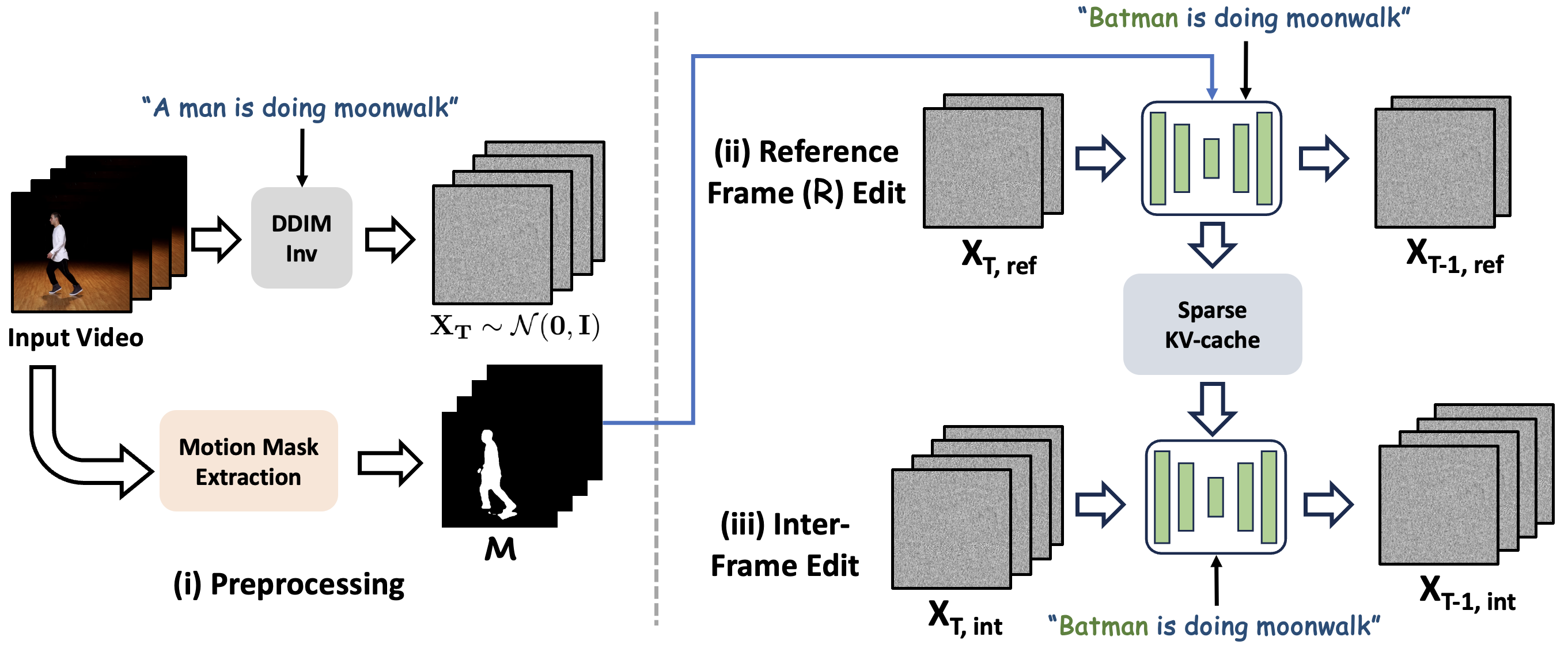}
\caption{Ada-VE Overview: (i) During preprocessing, DDIM inversion is performed to extract deterministic noise $\mathbf{X}_T \sim \mathcal{N}(0, I)$, and successive coarse motion masks $\mathcal{M}$ are extracted using a lightweight optical flow model, in a single step. (ii) Several reference frames $\mathbf{X}_{t, ref}$ are then sampled at timestep $t$, and jointly edited iteratively with the proposed sparse extension of self-attention KVs guided by motion masks $\mathbf{\mathbf{\mathcal{M}}}$, with all extended KVs being cached. (iii) Finally, all intermediate frames $\mathbf{X}_{t, int}$ are edited using the cached sparse reference KVs at timestep $t$.}
\label{fig:2}
\vspace{-2mm}
\end{figure*}

\subsection{Diffusion Probabilistic Models (DPM)}
DPMs~\cite{nichol2021glide, ddpm, croitoru2022diffusion, nichol2021improved} introduce a family of generative models that learn a data distribution through iterative progressive denoising of a Gaussian i.i.d noisy image $\mathbf{X}_T \sim \mathcal{N}(0, I)$. DPM generates a clean image ${x}_0$ of the target conditional distribution $q$ using additional text guidance $\mathcal{T}$.
Subsequent work \cite{rombach2022high, blattmann2023videoldm} introduced Latent Diffusion Models (LDMs) to generate high-resolution images by processing the conditional denoising in the latent space. There are multiple components in the LDM architecture, relying on the base U-Net architecture. Initially, a pre-trained encoder $\mathcal{E}$ and decoder $\mathcal{D}$ are trained to extract intermediate latent representations and generate representative images from the latents, respectively. The core denoising components consist of: a residual convolutional block to preserve spatial features, a self-attention (SA) block for feature recalibration, followed by a cross-attention block to introduce text guidance. In particular, SA blocks~\cite{vaswani2017attention} measure the feature affinity across projected $d$-dimensional query and key features $Q$ and $K$, respectively, followed by recalibration with value features $V$.
The self-attention plays a critical role in preserving the structures of the generated frames~\cite{pnpDiffusion2023, zhang2023controlvideo, geyer2023tokenflow}.

\subsection{Controlled Editing with LDM}
Song \textit{et al.}~\cite{song2020_ddim} introduced a deterministic sampling algorithm, DDIM, to sample intermediate noisy frames from the initial noise ${x}_T$. By using DDIM inversion, the deterministic noise ${x}_T \sim \mathcal{N}(0,I)$ can be extracted from a clean image ${x}_0$. Starting from deterministic noise with modified text prompts (SDEdit) is a viable approach for image editing~\cite{meng2022sdedit}. However, this process can lose some structural properties. Consequently, follow-up methods have focused on injecting structural priors from the given image. ControlNet~\cite{controlnet} introduced structurally encoded feature injections from extracted edge or depth information from the video. PnP~\cite{pnpDiffusion2023} proposed injecting self-attention and residual block features from guidance frames.

These zero-shot image editing techniques can be applied directly to videos. However, a major challenge is achieving consistent generation across frames. Prior work introduced extended cross-frame self-attention ~\cite{geyer2023tokenflow, peng2024conditionvideo} to preserve generated content by extending the keys and values of each $i^{th}$ frame of total $N$ frames in each self-attention using:
\begin{equation}
\begin{aligned}
& \mathtt{ExtendedSelfAttn}(X_i) = \mathtt{Softmax} \left(\frac{Q_i K_{\mathtt{all}}^T}{\sqrt{d}}\right) V_{\mathtt{all}}, \\
& \text{where} \ K_{\mathtt{all}} = \{ K_n \}_{n=1}^N, V_{\mathtt{all}} = \{ V_n \}_{n=1}^N
\end{aligned}
\end{equation}
While such extensions play a crucial role in consistent character generation, naively extending keys $K_{\mathtt{all}}$ and values $V_{\mathtt{all}}$ across all frames consumes significant memory and computational resources. Moreover, despite achieving consistent character generation, other challenges include generating smooth, temporally consistent frames over the video. 
\section{Methodology}
\label{method}
To increase the number of frames in joint editing and reduce computational burden, we propose a motion-adaptive KV selection method. This method derives the KV extension directly from the motion guidance of the given video. Additionally, to generate temporally consistent intermediate frames, we introduce the KV-cache technique, which reuses KVs from key pivotal frames during joint editing.
Our method builds on Plug-and-Play (PnP)~\cite{pnpDiffusion2023} image editing methods, which start with DDIM-inversion of the guidance video, followed by injecting  self-attention and residual features during final editing. Our primary contribution is the \textit{motion-adaptive} extension of the \textit{cross-frame self-attention} method, and the \textit{KV-caching technique} for enhanced video synthesis (see Fig.~\ref{fig:2}). The following sections present the proposed approach in detail.

\begin{figure*}[t]
\centering
\includegraphics[width= 0.85\textwidth]{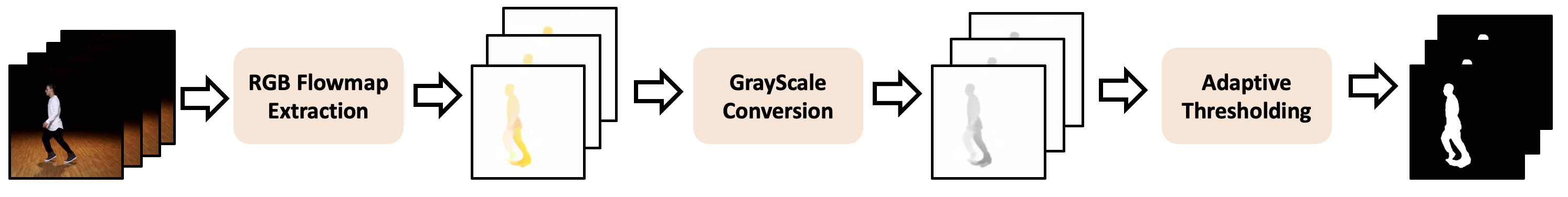}
\caption{\textbf{Motion Mask Extraction:} A lightweight, off-the-shelf model is employed to extract coarse optical flow maps from each pair of successive frames. These flow maps are converted to RGB and then to grayscale images. Finally, a thresholding technique is applied to extract moving region masks for each frame. This operation is performed once during preprocessing in a single step.}
\label{fig:3}
\vspace{-2mm}
\end{figure*}

\subsection{Adaptive Cross-Frame Attention}
Since in general video contains redundant information, we selectively perform KV-extension in self-attention based on temporal motion. Moving regions in each frame contain salient information about dynamic structures, whereas stationary regions generally contain repetitive features. Our primary goal is to capture more details from moving regions across multiple frames while sparsely integrating stationary features, thereby reducing the reliance on precise motion estimation. This process can be divided into two parts: extract motion masks from the guidance video using off-the-shelf optical flow models, and integrate these masks to selectively extend reference frame KVs in self-attention blocks. 

We note that motion mask extraction is performed using real-time, lightweight optical flow models to coarsely highlight moving regions. This extraction is only done once, while the T2I diffusion model runs iteratively over multiple timesteps (\textit{e.g.}, DDIM runs 50 steps), making the computational cost of flow estimation negligible by comparison. Additionally, our sparse integration of stationary regions reduces the need for precise estimation of moving regions with optical flow models, which is often impractical due to challenges such as occlusions, motion blur, and complex background movements.

\begin{figure}[t]
\centering
\includegraphics[width= 1.0\columnwidth]{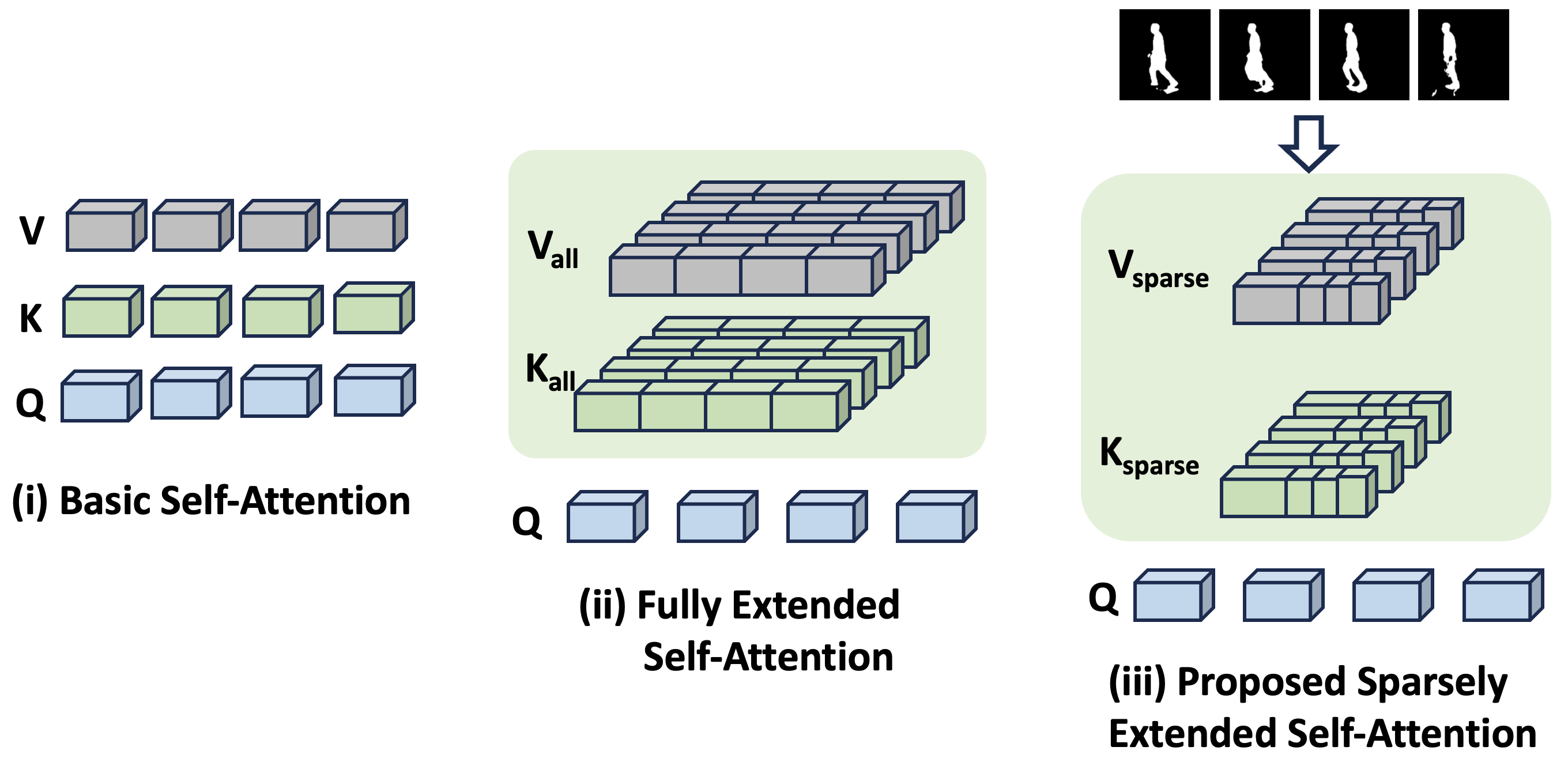}
\caption{(i) \textbf{Basic Self-Attention:} Queries (Q), keys (K), and values (V) are independently used for each frame. (ii) \textbf{Fully Extended Self-Attention:} All keys and values are combined into $K_{all}$ and $V_{all}$ for cross-frame self-attention. (iii) \textbf{Proposed Sparsely Extended Self-Attention:} Keys and values from all frames are sparsely extended into $K_{sparse}$ and $V_{sparse}$ to capture more details of moving regions than stationary background regions, utilizing motion masks $\mathbf{\mathcal{M}}$.}
\label{fig:4}
\vspace{-2mm}
\end{figure}

\subsubsection{Successive Motion Mask Extraction}
Following prior work~\cite{yang2023rerender, geyer2023tokenflow, zhang2023controlvideo, peng2024conditionvideo}, we initially sample a total of $Z$ reference frames $\mathbf{R} = \{R_1, R_2, \dots, R_Z\}$ from a total of $N$ frames for joint-editing. The sampling interval $s$ can be adjusted based on the video length and computational overhead. Instead of na\"ively extending cross-frame attention across $Z$ reference frames with inflated T2I diffusion models~\cite{zhang2023controlvideo, geyer2023tokenflow}, we propose adaptive sparse extension of KV features to reduce redundant computations. We extract successive motion masks $M_i$ from these $Z$ frames to separate moving regions from stationary regions, which are later used for sparse extended self-attention operations (see Fig.~\ref{fig:3}).

We extract optical flow maps $F_{i} \in \mathbb{R}^{h \times w \times 2}$ between each pair of reference frames $R_{i-1} \rightarrow R_{i}$ where $R_i \in \mathbb{R}^{h \times w \times 3}$ using an off-the-shelf flow estimator (FlowNet-v2~\cite{ilg2017flownet}). Afterwards, the flow field $F_{i}$ is converted to RGB images and subsequently, to gray-scale images $G_{i} \in \mathbb{R}^{h \times w}$ using basic image processing tools that highlight the moving regions in frame $R_{i}$ with respect to frame $R_{i-1}$, given by:
\begin{equation}
\begin{aligned}
    F_{i} &= \mathtt{FlowEst}(R_{i-1} \rightarrow R_{i}), \quad \forall i \in \{2, 3, \dots, Z \}, \\
    G_{i} &= \mathtt{RGB2Gray} \left( \mathtt{Flow2RGB}(F_{i}) \right)
\end{aligned}
\end{equation}

Afterwards, stored gray-scale optical flow visualizations $G_{i}$ of each reference frame are used in every self-attention block for sparse sampling of KV features. Each self-attention block operates at a smaller spatial resolution than $(h \times w)$. However, LDM preserves location features throughout the network, facilitating the reuse of spatial motion priors encoded in $G_i$. Hence, we perform downsampling (DS) on $G_{i}$ to match the spatial resolution $(d^j \times d^j)$ of the feature map $X^j$. A simple adaptive thresholding operation (\textit{e.g.}, Otsu's thresholding~\cite{otsu1975threshold}) extracts spatial moving region masks $M^j$, given by:
\begin{equation}
    M_{i}^j = \mathtt{ApplyThr}(\mathtt{DW}(G_{i}, \tau)), \quad \forall i = \{2, \dots, Z \}
\end{equation}
where $\tau = d^j / h$ represents the downsampling ratio at the $j^{th}$ self-attention block. These motion masks are used in the sparse extension of joint self-attention of reference frames, in every diffusion sampling iteration.

\subsubsection{Adaptive Extension of Self-Attention}
The key idea of adaptive extension of self-attention is to integrate more features of the moving regions while eliminating redundant features from the stationary ones. We use estimated motion masks $M_{i}^j$ for aggregating moving region features at every $j^{th}$ self-attention block. However, successive motion masks may lose some key features, especially at the surroundings due to occlusions, motion blur, and background movements.
To overcome this, we integrate full-frame KV features at a sparser frame interval $r$ ($r >> s$) that overcomes the feature loss. Moreover, we only use the moving region features from most other reference frames based on motion mask $M$ eliminating redundant computations (see Fig.~\ref{fig:4}). The proposed extended sparse KV can be represented as:

\begin{equation}
\begin{aligned}
    \mathbf{K}_{\mathtt{sparse}}^j &= \left\{ K_1^j, \dots, (M_{r-1}^j \cdot K_{r-1}^j), K_r^j, \right. \\
    &\quad \left. (M_{r+1}^j \cdot K_{r+1}^j), \dots, K_{2r}^j, \dots, K_Z^j \right\}, \\
    \mathbf{V}_{\mathtt{sparse}}^j &= \left\{ V_1^j, \dots, (M_{r-1}^j \cdot V_{r-1}^j), V_r^j, \right. \\
    &\quad \left. (M_{r+1}^j \cdot V_{r+1}^j), \dots, V_{2r}^j, \dots, V_Z^j \right\}
\end{aligned}    
\end{equation}

where $M \cdot K$ represents sparsely sampled key features using motion mask $M$. The proposed sparse extension of self-attention, $\mathtt{SESA}(\cdot)$, at the $i^{th}$ block is given by:
\begin{equation}
    \mathtt{SESA}(X_i) = \mathtt{Softmax} \left(\frac{Q_i K_{\mathtt{sparse}}^T}{\sqrt{d}}\right) V_{\mathtt{sparse}}, I_i \in \mathbf{R}    
\end{equation}
The proposed KV-extension in $\mathtt{SESA}(\cdot)$ selectively retains moving region tokens from all frames. Hence, slower moving regions will adaptively incorporate sparser sampling of KVs, while faster moving regions will leverage denser sampling of KVs across all sampled reference frames $\mathbf{R}$.

\subsection{KV-Caching and Inter-Frame Editing}
Operating extended joint self-attention across all video frames is computationally intensive. To address this, we employ sparser sampling rates of reference frames (higher $s$, $r$) in the initial phase, and perform joint editing. Subsequently, we generate intermediate frames in the second phase leveraging KV features stored from the initial phase. Previous approaches often rely on frame interpolation techniques for intermediate frames~\cite{yang2023rerender, liang2023flowvid}, but these methods can result in blurring effects or inconsistent character generation. Instead of interpolation, Ada-VE focuses on directly generating intermediate frames with consistent characters to better preserve visual quality.

Our empirical observations indicate that maintaining a consistent KV feature space is crucial for generating consistent characters across intermediate frames. Rather than using only nearby KVs for intermediate frame generation~\cite{zhang2023controlvideo}, we propose caching the sparsely sampled KVs from all reference frames (\textit{i.e.}, $\mathbf{K}_{\mathtt{sparse}}$, $\mathbf{V}_{\mathtt{sparse}}$) during the first pass. These cached KVs are then reused with the queries $Q_m$ from all intermediate frames $I_m \notin \mathbf{R}$ in self-attention ($\mathtt{IFSA}(\cdot)$) during the second pass:
\begin{equation}
    \mathtt{IFSA}(X_m) = \mathtt{Softmax} \left(\frac{Q_m K_{\mathtt{sparse}}^T}{\sqrt{d}}\right) V_{\mathtt{sparse}}
\end{equation}
The proposed $\mathtt{IFSA}(\cdot)$ helps in generating consistent characters in intermediate frames without compromising visual quality or further extending the joint feature space. Additionally, because self-attention mechanisms are responsible for maintaining structural priors, which can lead to subtle motion inconsistencies across frames, we adopt a hierarchical sampling technique from~\cite{zhang2023controlvideo}. This technique ensures smooth motion transitions by interleaving the generated frames with interpolated frames throughout the video.
\section{Results}
\label{results}
\subsection{Main results}

\begin{figure*}[t]
\centering
\includegraphics[width= 0.8\textwidth]{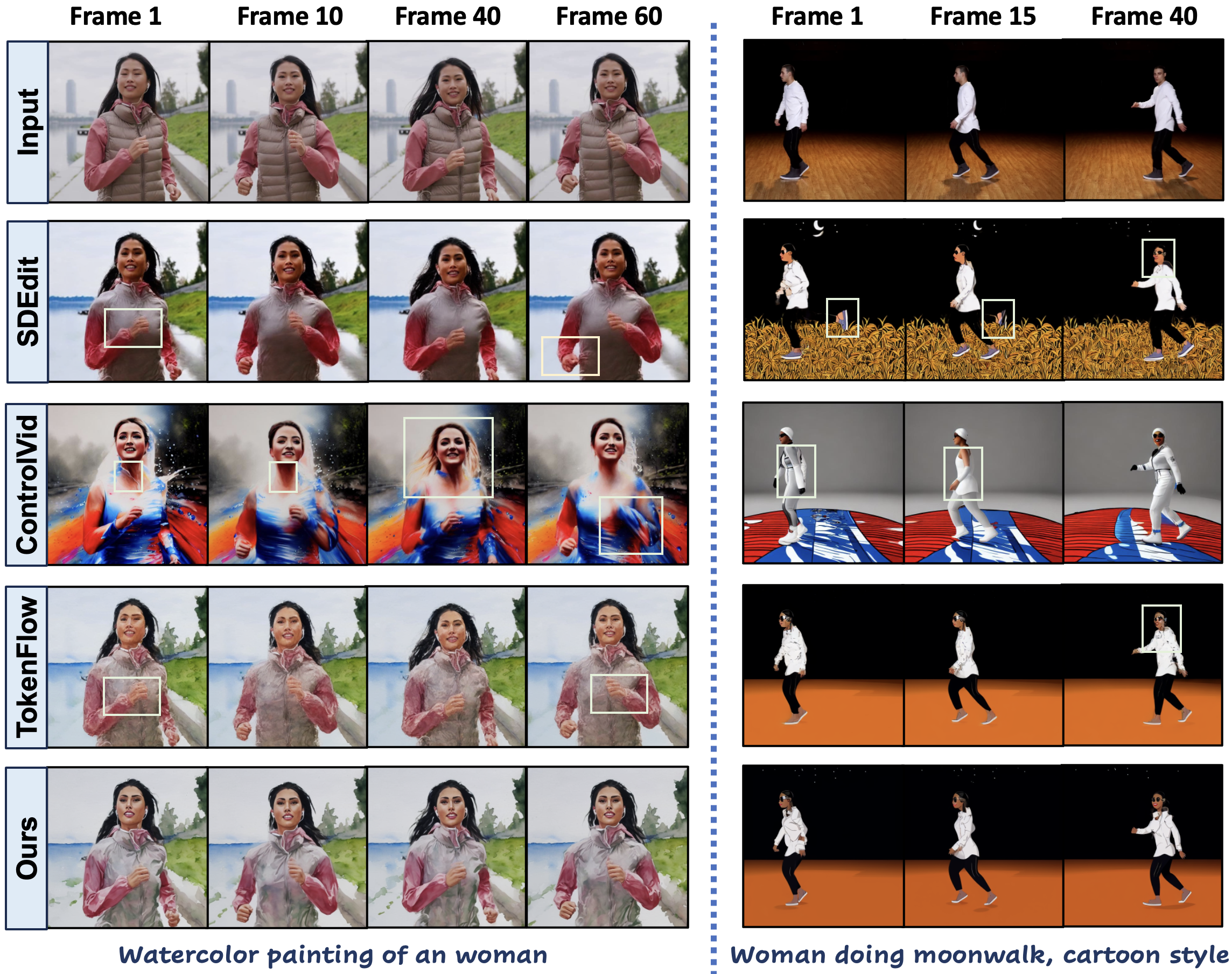}
\caption{Qualitative comparisons with state-of-the-art video editing methods show that SDEdit~\cite{meng2022sdedit} struggles with motion blur and inconsistent character generation in longer videos, while ControlVideo~\cite{zhang2023controlvideo} also has inconsistency issues. TokenFlow~\cite{geyer2023tokenflow} offers better temporal consistency but still suffers from motion blur. In contrast, Ada-VE (ours) achieves superior visual quality and consistency throughout, building on the PnP~\cite{pnpDiffusion2023} image editing baseline like TokenFlow but delivering higher-quality results.}
\label{fig:5}
\vspace{-2mm}
\end{figure*}

\paragraph{User preference study:} 
We used 15 full-length DAVIS~\cite{davis} videos ranging from 40 to 200 frames, and 50 manually designed prompts to conduct quantitative evaluations based on user preference. We had 10 participants who were selected through voluntary participation. All participants were fully informed about the nature of the study.
 Users were asked to select the best-edited videos in three categories: 
(1) \textbf{Visual Quality}: This considers the amount of blurriness and overall visual perception of users on edited videos; 
(2) \textbf{Temporal Consistency}: This represents the temporal motion smoothness of edited videos; and 
(3) \textbf{Content Consistency}: This denotes the generation of consistent characters across all frames. 

Six baseline methods were used for comparative analysis: Text2Video-zero~\cite{text2video-zero}, SDEdit~\cite{meng2022sdedit}, TokenFlow~\cite{geyer2023tokenflow}, ControlVideo~\cite{zhang2023controlvideo}, Motion-Transfer~\cite{motion_transfer}, and StreamV2V~\cite{streamv2v}. Note that only 16 frames of generated videos were considered for Text2Video-zero and MotionTransfer due to extensive memory requirements for longer videos. Moreover, we considered TokenFlow integrated SDEdit~\cite{meng2022sdedit} to make it a stronger video baseline for a fair comparison.

\begin{table}[t]
    \centering
    \caption{User preference study is conducted on visual quality, temporal consistency and content consistency with the same prompts. * denotes shorter video length for memory requirements.}
    \label{t1}
    \scalebox{0.8}{
    \begin{tabular}{cccc}
    \toprule
    \multirow{2}{*}{\textbf{Method}} & \multirow{2}{*}{\textbf{\begin{tabular}[c]{@{}c@{}}Visual\\ Quality (\%)\end{tabular}}} & \multirow{2}{*}{\textbf{\begin{tabular}[c]{@{}c@{}}Temporal \\ Cons.(\%)\end{tabular}}} & \multirow{2}{*}{\textbf{\begin{tabular}[c]{@{}c@{}}Content \\ Cons.(\%)\end{tabular}}} \\
                                     &                                                                                    &                                                                                     &                                                                                    \\    
    \midrule
    T2V-zero*~\cite{text2video-zero}       & 1.3                     & 1.1                    & 0.5                   \\
    MotionTransfer*~\cite{motion_transfer}       & 12.2                    & 15.3                    & 14.2                   \\
    SDEdit~\cite{meng2022sdedit}          & 10.1                     & 12.5                    & 10.3                   \\
ControlVideo~\cite{zhang2023controlvideo}    & 8.4                    & 4.6                    & 4.4                   \\
    StreamV2V~\cite{streamv2v}       & 8.1                    & 3.5                   & 2.8                   \\
    TokenFlow~\cite{geyer2023tokenflow}       & 21.3                    & \textbf{32.3}                    & 28.5                   \\
    Ada-VE(ours)     & \textbf{38.6}                    & 31.2                    & \textbf{39.3}                 \\ 
    \bottomrule
    \end{tabular}}
\vspace{-2mm}
\end{table}

The results are presented in Table~\ref{t1}. Ada-VE achieved the highest user preference in visual quality at 38.6\%, enhancing visual representation with reduced blurriness, and the highest content consistency at 39.3\%, generating consistent content across frames. In contrast, most other video synthesis methods struggled to generate consistent content and often exhibited temporal blurriness. Moreover, Ada-VE performed on par with the state-of-the-art TokenFlow method in terms of temporal consistency, while significantly outperforming other baseline methods. Hence, Ada-VE significantly improves visual quality and content consistency without sacrificing motion smoothness.

\vspace{-3mm}
\paragraph{Qualitative comparisons:}
We present a qualitative analysis of the baseline methods in Fig.~\ref{fig:5}. We highlight three strong baselines for this analysis with the same length of video: SDEdit~\cite{meng2022sdedit}, TokenFlow~\cite{geyer2023tokenflow}, and ControlVideo~\cite{zhang2023controlvideo}. In the example of the woman running, SDEdit and TokenFlow often produce blurry fingers and hands, missing important details. Additionally, SDEdit generates inconsistent legs and distorted face structures in the moonwalk videos. TokenFlow exhibits similar face distortions. ControlVideo generates inconsistent hairstyles and hands, with more prominent inter-frame inconsistencies in the moonwalk videos.  In contrast, Ada-VE generates consistent fingers in the woman running example with significantly reduced motion blurriness. Moreover, the face distortion in the moonwalk videos is notably improved in Ada-VE. These results demonstrate the superior quality of Ada-VE compared to state-of-the-art methods. Additionally, we provide extensive qualitative videos in the supplementary with all six baselines.

\begin{figure}[t]
    \centering
    \includegraphics[width=0.95\linewidth]{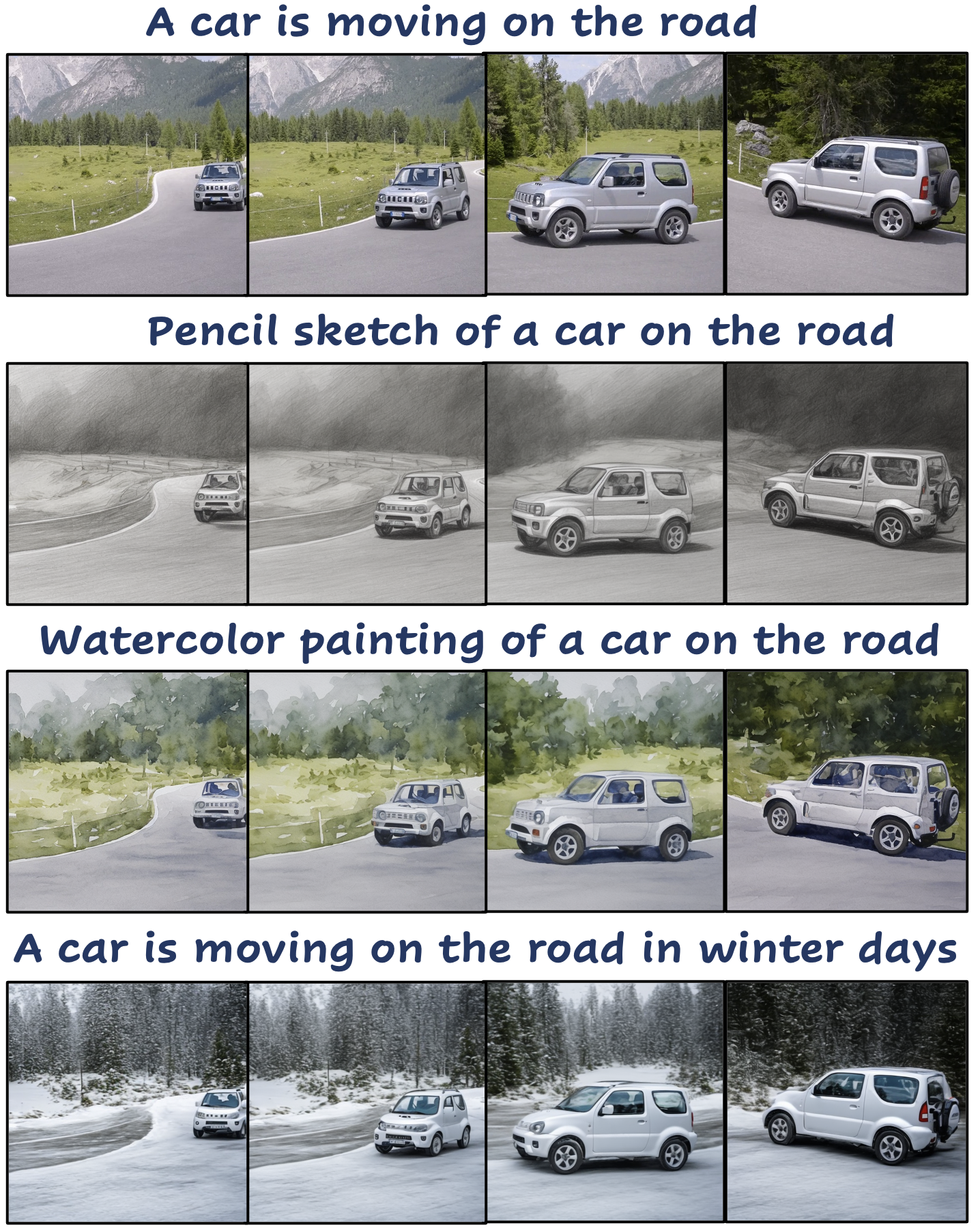}
        \caption{We present ablation study on video editing and style transfer prompts with proposed Ada-VE framework.}
    \label{fig:6}
\end{figure}

\begin{table}[t]
    \centering
    \caption{We present comparisons across various joint editing choices across 40 frames. Here, $s$ denotes the sampling interval for sparse extension. Denser extensions of self-attentions significantly improve the performance, whereas sparser sampling reduces latency. Our adaptive extension of self-attention can be very effective in reducing the latency without compromising performance, particularly in denser extensions.}
    \label{t2}
    \scalebox{0.8}{
    \begin{tabular}{ccccc}
    \toprule
    \multirow{2}{*}{\textbf{Method}}                                                & \multirow{2}{*}{\textbf{\begin{tabular}[c]{@{}c@{}}with\\ AdaVE\end{tabular}}} & \multirow{2}{*}{\textbf{\begin{tabular}[c]{@{}c@{}}Warp\\ err ($\downarrow$)\end{tabular}}} & \multirow{2}{*}{\textbf{\begin{tabular}[c]{@{}c@{}}CLIP\\ Score ($\uparrow$)\end{tabular}}} & \multirow{2}{*}{\textbf{Latency ($\downarrow$)}} \\
    
                                                                                    &                                                                                &                                                                              &                                                                                &                                   \\
    \midrule
    LDM-recons                                                                      & -                                                                              & 2.2                                                                          & 0.20                                                                           & -                                 \\
    \midrule
    First Only                                                                      & \xmark                                                                              & 37.4                                                                         & 0.25                                                                           & 1$\times$                                \\
    First+Prev                                                                      & \xmark                                                                              & 25.4                                                                         & 0.29                                                                           & 1.4$\times$                              \\
    First + TwoPrev                                                                 & \xmark                                                                              & 21.5                                                                         & 0.31                                                                           & 1.8$\times$                                \\
    \midrule
    \multirow{2}{*}{\begin{tabular}[c]{@{}c@{}}Sparse Sample \\ (\textit{s}=8)\end{tabular}} & \xmark                                                                              & 10.6                                                                         & 0.32                                                                           & 2.8$\times$                              \\
                                                                                    & \cmark                                                                              & 10.7                                                                         & 0.32                                                                           & 1.7$\times$                              \\
    \midrule
    \multirow{2}{*}{\begin{tabular}[c]{@{}c@{}}Sparse Sample \\ (\textit{s}=3)\end{tabular}} & \xmark                                                                              & 4.8                                                                          & 0.34                                                                           & 5.3$\times$                             \\
                                                                                    & \cmark                                                                              & 4.7                                                                          & 0.34                                                                           & 1.9$\times$                              \\
    \midrule
    \multirow{2}{*}{\begin{tabular}[c]{@{}c@{}}Fully\\ Extended (\textit{s}=1)\end{tabular}} & \xmark                                                                              & 2.8                                                                          & 0.35                                                                           & 13.6$\times$                             \\
                                                                                    & \cmark                                                                              & 2.8                                                                          & 0.35                                                                           & 3.5$\times$                            \\
        \bottomrule
    \end{tabular}}
\vspace{-2mm}
\end{table}

\vspace{-1mm}
\subsection{Ablation Study}
\vspace{-1mm}
\paragraph{Analysis joint editing performance:}
We study the effects of various extensions of self-attention performance in joint editing, as shown in Table~\ref{t2}. We selected 40 consecutive frames from 10 DAVIS videos for joint editing based on text prompts. To quantitatively measure temporal consistency and realistic prompt alignment, we used the warping-error and CLIP-score metrics following prior work~\cite{geyer2023tokenflow, zhang2023controlvideo}. Apart from the various self-attention extensions, the same baseline PnP model~\cite{pnpDiffusion2023} was used for this study.
Our observations based on Tab.~\ref{t2} indicate that increasing the number of reference frames directly enhances performance, although it significantly increases computational burden and latency. 
Incorporating our proposed adaptive extension of self-attention can significantly reduce latency without compromising performance. For instance, with full extensions of self-attention, Ada-VE can achieve 3.9$\times$ speed-up on average. We use single A6000 GPUs with 48GB memory for experiments.
In other words,while using the same GPU memory, proposed sparse extension facilitates use of average $3\times$ higher reference frames than baseline full extension, with $25\%$ higher CLIP score and $45\%$ lower warp error (Tab.~\ref{n1}).

\begin{table}[t]
\centering
\caption{With the same GPU memory, Ada-VE can operate with more reference frames contributing to performance improvement.}
\label{n1}
\scalebox{0.8}{
\begin{tabular}{cccccc}
\toprule
\multirow{2}{*}{\textbf{Extension}} & \multirow{2}{*}{\textbf{\begin{tabular}[c]{@{}c@{}}\# Ref. \\ Frames $\uparrow$\end{tabular}}} & \multirow{2}{*}{\textbf{\begin{tabular}[c]{@{}c@{}}GPU\\ Mem $\downarrow$ \end{tabular}}} & \multirow{2}{*}{\textbf{Latency $\downarrow$}} & \multirow{2}{*}{\textbf{\begin{tabular}[c]{@{}c@{}}CLIP\\ Score $\uparrow$\end{tabular}}} & \multirow{2}{*}{\textbf{\begin{tabular}[c]{@{}c@{}}Warp\\ Err $\downarrow$\end{tabular}}} \\

                           &                                                                       &                                                                    &                          &                                                                       &                                                                     \\
\midrule
Full Ext.                       & 20                                                                    & 1x                                                                 & 1x                       & 0.31                                                                  & 3.8                                                                 \\
Prop. Ext.                      & \textbf{61}                                                                    & 1x                                                                 & \textbf{0.8x}                     & \textbf{0.39}                                                          & \textbf{2.1}                                                       \\
\bottomrule
\end{tabular}}
\vspace{-3mm}
\end{table}

\vspace{-2mm}
\paragraph{Qualitative analysis:} We present a qualitative ablation study on style transfer and video editing prompts in Fig.~\ref{fig:6}. We perform various prompts, such as pencil sketch, watercolor painting, and atmosphere switching. Ada-VE consistently modifies the surroundings and overall style while preserving the details of the moving subject. As the examples show, Ada-VE can modify the subject while preserving visual aesthetics and sharp details throughout the video.

\vspace{-1mm}
\section{Limitations}
\vspace{-1mm}
\label{limitations}
Despite its strengths, Ada-VE inherits some limitations from the underlying Plug-and-Play diffusion models, particularly in handling complete structural alterations of the guidance video. However, the techniques introduced in this work are widely applicable and can be integrated into most existing image and video editing frameworks.

\vspace{-1mm}
\section{Conclusion}
\vspace{-1mm}
In this paper, we presented Ada-VE, a novel video-to-video synthesis method designed to overcome the limitations of existing approaches. By introducing an adaptive motion-based KV selection strategy and KV-caching, Ada-VE significantly reduces computational overhead while improving visual quality and content consistency. Leveraging sparse KV feature extraction within Plug-and-Play diffusion models, our method achieves superior performance with a lower computational cost.
Through extensive qualitative and quantitative evaluations, Ada-VE demonstrated substantial improvements over state-of-the-art methods, particularly in maintaining visual clarity, temporal consistency, and content fidelity across frames. Its ability to modify moving subjects consistently while preserving fine details was validated in various challenging video editing tasks.
Ada-VE represents a strong advancement in video synthesis, providing a practical solution for generating high-quality, temporally consistent videos and opening new directions for future research in video editing and synthesis.

\section*{Acknowledgement}
This project is supported by the ONR Minerva program, iMAGiNE - the Intelligent Machine Engineering Consortium at UT Austin, and a UT Cockrell School of Engineering Doctoral Fellowship.

{\small
\bibliographystyle{ieee_fullname}
\bibliography{reference}

\begin{thebibliography}{10}\itemsep=-1pt

\bibitem{bar2022text2live}
Omer Bar-Tal, Dolev Ofri-Amar, Rafail Fridman, Yoni Kasten, and Tali Dekel.
\newblock Text2live: Text-driven layered image and video editing.
\newblock In {\em European Conference on Computer Vision}, pages 707--723. Springer, 2022.

\bibitem{bar2023multidiffusion}
Omer Bar-Tal, Lior Yariv, Yaron Lipman, and Tali Dekel.
\newblock Multidiffusion: Fusing diffusion paths for controlled image generation.
\newblock {\em arXiv preprint arXiv:2302.08113}, 2023.

\bibitem{blattmann2023videoldm}
Andreas Blattmann, Robin Rombach, Huan Ling, Tim Dockhorn, Seung~Wook Kim, Sanja Fidler, and Karsten Kreis.
\newblock Align your latents: High-resolution video synthesis with latent diffusion models.
\newblock In {\em IEEE Conference on Computer Vision and Pattern Recognition ({CVPR})}, 2023.

\bibitem{cao2023masactrl}
Mingdeng Cao, Xintao Wang, Zhongang Qi, Ying Shan, Xiaohu Qie, and Yinqiang Zheng.
\newblock Masactrl: Tuning-free mutual self-attention control for consistent image synthesis and editing, 2023.

\bibitem{Ceylan2023Pix2VideoVE}
Duygu Ceylan, Chun-Hao~Paul Huang, and Niloy~Jyoti Mitra.
\newblock Pix2video: Video editing using image diffusion.
\newblock {\em ArXiv}, abs/2303.12688, 2023.

\bibitem{chefer2023attend}
Hila Chefer, Yuval Alaluf, Yael Vinker, Lior Wolf, and Daniel Cohen-Or.
\newblock Attend-and-excite: Attention-based semantic guidance for text-to-image diffusion models.
\newblock {\em arXiv preprint arXiv:2301.13826}, 2023.

\bibitem{video_epitome}
V. Cheung, B.J. Frey, and N. Jojic.
\newblock Video epitomes.
\newblock In {\em 2005 IEEE Computer Society Conference on Computer Vision and Pattern Recognition (CVPR'05)}, 2005.

\bibitem{croitoru2022diffusion}
Florinel-Alin Croitoru, Vlad Hondru, Radu~Tudor Ionescu, and Mubarak Shah.
\newblock Diffusion models in vision: A survey.
\newblock {\em arXiv preprint arXiv:2209.04747}, 2022.

\bibitem{beatgan}
Prafulla Dhariwal and Alexander Nichol.
\newblock Diffusion models beat gans on image synthesis.
\newblock {\em Advances in Neural Information Processing Systems}, 2021.

\bibitem{gen1}
Patrick Esser, Johnathan Chiu, Parmida Atighehchian, Jonathan Granskog, and Anastasis Germanidis.
\newblock Structure and content-guided video synthesis with diffusion models.
\newblock {\em arXiv preprint arXiv:2302.03011}, 2023.

\bibitem{geyer2023tokenflow}
Michal Geyer, Omer Bar-Tal, Shai Bagon, and Tali Dekel.
\newblock Tokenflow: Consistent diffusion features for consistent video editing.
\newblock {\em arXiv preprint arXiv:2307.10373}, 2023.

\bibitem{gupta2022rv}
Sonam Gupta, Arti Keshari, and Sukhendu Das.
\newblock Rv-gan: Recurrent gan for unconditional video generation.
\newblock In {\em Proceedings of the IEEE/CVF Conference on Computer Vision and Pattern Recognition}, pages 2024--2033, 2022.

\bibitem{ho2022imagen_video}
Jonathan Ho, William Chan, Chitwan Saharia, Jay Whang, Ruiqi Gao, Alexey Gritsenko, Diederik~P Kingma, Ben Poole, Mohammad Norouzi, David~J Fleet, et~al.
\newblock Imagen video: High definition video generation with diffusion models.
\newblock {\em arXiv preprint arXiv:2210.02303}, 2022.

\bibitem{ddpm}
Jonathan Ho, Ajay Jain, and Pieter Abbeel.
\newblock Denoising diffusion probabilistic models.
\newblock {\em Advances in Neural Information Processing Systems}, 2020.

\bibitem{hong2022improving}
Susung Hong, Gyuseong Lee, Wooseok Jang, and Seungryong Kim.
\newblock Improving sample quality of diffusion models using self-attention guidance.
\newblock {\em arXiv preprint arXiv:2210.00939}, 2022.

\bibitem{hui2018liteflownet}
Tak-Wai Hui, Xiaoou Tang, and Chen~Change Loy.
\newblock Liteflownet: A lightweight convolutional neural network for optical flow estimation.
\newblock In {\em Proceedings of the IEEE conference on computer vision and pattern recognition}, pages 8981--8989, 2018.

\bibitem{ilg2017flownet}
Eddy Ilg, Nikolaus Mayer, Tonmoy Saikia, Margret Keuper, Alexey Dosovitskiy, and Thomas Brox.
\newblock Flownet 2.0: Evolution of optical flow estimation with deep networks.
\newblock In {\em Proceedings of the IEEE conference on computer vision and pattern recognition}, pages 2462--2470, 2017.

\bibitem{text2video-zero}
Levon Khachatryan, Andranik Movsisyan, Vahram Tadevosyan, Roberto Henschel, Zhangyang Wang, Shant Navasardyan, and Humphrey Shi.
\newblock Text2video-zero: Text-to-image diffusion models are zero-shot video generators.
\newblock {\em arXiv preprint arXiv:2303.13439}, 2023.

\bibitem{Khachatryan2023Text2VideoZeroTD}
Levon Khachatryan, Andranik Movsisyan, Vahram Tadevosyan, Roberto Henschel, Zhangyang Wang, Shant Navasardyan, and Humphrey Shi.
\newblock Text2video-zero: Text-to-image diffusion models are zero-shot video generators.
\newblock {\em ArXiv}, abs/2303.13439, 2023.

\bibitem{streamdiff}
Akio Kodaira, Chenfeng Xu, Toshiki Hazama, Takanori Yoshimoto, Kohei Ohno, Shogo Mitsuhori, Soichi Sugano, Hanying Cho, Zhijian Liu, and Kurt Keutzer.
\newblock Streamdiffusion: A pipeline-level solution for real-time interactive generation.
\newblock {\em arXiv preprint arXiv:2312.12491}, 2023.

\bibitem{lee2023textvideoedit}
Yao-Chih Lee, Ji-Ze Genevieve~Jang Jang, Yi-Ting Chen, Elizabeth Qiu, and Jia-Bin Huang.
\newblock Shape-aware text-driven layered video editing demo.
\newblock {\em arXiv preprint arXiv:2301.13173}, 2023.

\bibitem{streamv2v}
Feng Liang, Akio Kodaira, Chenfeng Xu, Masayoshi Tomizuka, Kurt Keutzer, and Diana Marculescu.
\newblock Looking backward: Streaming video-to-video translation with feature banks.
\newblock {\em arXiv preprint arXiv:2405.15757}, 2024.

\bibitem{liang2023flowvid}
Feng Liang, Bichen Wu, Jialiang Wang, Licheng Yu, Kunpeng Li, Yinan Zhao, Ishan Misra, Jia-Bin Huang, Peizhao Zhang, Peter Vajda, et~al.
\newblock Flowvid: Taming imperfect optical flows for consistent video-to-video synthesis.
\newblock {\em arXiv preprint arXiv:2312.17681}, 2023.

\bibitem{Liu2023VideoP2PVE}
Shaoteng Liu, Yuecheng Zhang, Wenbo Li, Zhe Lin, and Jiaya Jia.
\newblock Video-p2p: Video editing with cross-attention control.
\newblock {\em ArXiv}, abs/2303.04761, 2023.

\bibitem{liu2024sora}
Yixin Liu, Kai Zhang, Yuan Li, Zhiling Yan, Chujie Gao, Ruoxi Chen, Zhengqing Yuan, Yue Huang, Hanchi Sun, Jianfeng Gao, et~al.
\newblock Sora: A review on background, technology, limitations, and opportunities of large vision models.
\newblock {\em arXiv preprint arXiv:2402.17177}, 2024.

\bibitem{ma2023directed}
Wan-Duo~Kurt Ma, JP Lewis, W~Bastiaan Kleijn, and Thomas Leung.
\newblock Directed diffusion: Direct control of object placement through attention guidance.
\newblock {\em arXiv preprint arXiv:2302.13153}, 2023.

\bibitem{meng2022sdedit}
Chenlin Meng, Yutong He, Yang Song, Jiaming Song, Jiajun Wu, Jun-Yan Zhu, and Stefano Ermon.
\newblock {SDE}dit: Guided image synthesis and editing with stochastic differential equations.
\newblock In {\em International Conference on Learning Representations}, 2022.

\bibitem{munoz2021temporal}
Andres Munoz, Mohammadreza Zolfaghari, Max Argus, and Thomas Brox.
\newblock Temporal shift gan for large scale video generation.
\newblock In {\em Proceedings of the IEEE/CVF Winter Conference on Applications of Computer Vision}, pages 3179--3188, 2021.

\bibitem{nichol2021glide}
Alex Nichol, Prafulla Dhariwal, Aditya Ramesh, Pranav Shyam, Pamela Mishkin, Bob McGrew, Ilya Sutskever, and Mark Chen.
\newblock Glide: Towards photorealistic image generation and editing with text-guided diffusion models.
\newblock {\em arXiv preprint arXiv:2112.10741}, 2021.

\bibitem{nichol2021improved}
Alexander~Quinn Nichol and Prafulla Dhariwal.
\newblock Improved denoising diffusion probabilistic models.
\newblock In {\em International Conference on Machine Learning}, pages 8162--8171. PMLR, 2021.

\bibitem{otsu1975threshold}
Nobuyuki Otsu et~al.
\newblock A threshold selection method from gray-level histograms.
\newblock {\em Automatica}, 11(285-296):23--27, 1975.

\bibitem{peng2024conditionvideo}
Bo Peng, Xinyuan Chen, Yaohui Wang, Chaochao Lu, and Yu Qiao.
\newblock Conditionvideo: Training-free condition-guided video generation.
\newblock In {\em Proceedings of the AAAI Conference on Artificial Intelligence}, volume~38, pages 4459--4467, 2024.

\bibitem{davis}
Jordi Pont-Tuset, Federico Perazzi, Sergi Caelles, Pablo Arbel{\'a}ez, Alex Sorkine-Hornung, and Luc Van~Gool.
\newblock The 2017 davis challenge on video object segmentation.
\newblock {\em arXiv preprint arXiv:1704.00675}, 2017.

\bibitem{qi2023fatezero}
Chenyang Qi, Xiaodong Cun, Yong Zhang, Chenyang Lei, Xintao Wang, Ying Shan, and Qifeng Chen.
\newblock Fatezero: Fusing attentions for zero-shot text-based video editing.
\newblock {\em arXiv:2303.09535}, 2023.

\bibitem{rombach2022high}
Robin Rombach, Andreas Blattmann, Dominik Lorenz, Patrick Esser, and Bj{\"o}rn Ommer.
\newblock High-resolution image synthesis with latent diffusion models.
\newblock In {\em Proceedings of the IEEE/CVF Conference on Computer Vision and Pattern Recognition}, pages 10684--10695, 2022.

\bibitem{vid_style_transfer}
Manuel Ruder, Alexey Dosovitskiy, and Thomas Brox.
\newblock Artistic style transfer for videos.
\newblock In {\em Pattern Recognition - 38th German Conference (GCPR)}, 2016.

\bibitem{make_a_video}
Uriel Singer, Adam Polyak, Thomas Hayes, Xi Yin, Jie An, Songyang Zhang, Qiyuan Hu, Harry Yang, Oron Ashual, Oran Gafni, Devi Parikh, Sonal Gupta, and Yaniv Taigman.
\newblock Make-a-video: Text-to-video generation without text-video data, 2022.

\bibitem{song2020_ddim}
Jiaming Song, Chenlin Meng, and Stefano Ermon.
\newblock Denoising diffusion implicit models.
\newblock In {\em International Conference on Learning Representations}, 2020.

\bibitem{teed2020raft}
Zachary Teed and Jia Deng.
\newblock Raft: Recurrent all-pairs field transforms for optical flow.
\newblock In {\em Computer Vision--ECCV 2020: 16th European Conference, Glasgow, UK, August 23--28, 2020, Proceedings, Part II 16}. Springer, 2020.

\bibitem{pnpDiffusion2023}
Narek Tumanyan, Michal Geyer, Shai Bagon, and Tali Dekel.
\newblock Plug-and-play diffusion features for text-driven image-to-image translation.
\newblock {\em Proceedings of the IEEE/CVF Conference on Computer Vision and Pattern Recognition (CVPR)}, June 2023.

\bibitem{vaswani2017attention}
Ashish Vaswani, Noam Shazeer, Niki Parmar, Jakob Uszkoreit, Llion Jones, Aidan~N Gomez, {\L}ukasz Kaiser, and Illia Polosukhin.
\newblock Attention is all you need.
\newblock {\em Advances in neural information processing systems}, 30, 2017.

\bibitem{wang2020imaginator}
Yaohui Wang, Piotr Bilinski, Francois Bremond, and Antitza Dantcheva.
\newblock Imaginator: Conditional spatio-temporal gan for video generation.
\newblock In {\em Proceedings of the IEEE/CVF Winter Conference on Applications of Computer Vision}, pages 1160--1169, 2020.

\bibitem{wu2022tuneavideo}
Jay~Zhangjie Wu, Yixiao Ge, Xintao Wang, Stan~Weixian Lei, Yuchao Gu, Wynne Hsu, Ying Shan, Xiaohu Qie, and Mike~Zheng Shou.
\newblock Tune-a-video: One-shot tuning of image diffusion models for text-to-video generation.
\newblock {\em arXiv preprint arXiv:2212.11565}, 2022.

\bibitem{yang2023rerender}
Shuai Yang, Yifan Zhou, Ziwei Liu, and Chen~Change Loy.
\newblock Rerender a video: Zero-shot text-guided video-to-video translation, 2023.

\bibitem{motion_transfer}
Danah Yatim, Rafail Fridman, Omer Bar-Tal, Yoni Kasten, and Tali Dekel.
\newblock Space-time diffusion features for zero-shot text-driven motion transfer.
\newblock In {\em Proceedings of the IEEE/CVF Conference on Computer Vision and Pattern Recognition}, pages 8466--8476, 2024.

\bibitem{controlnet}
Lvmin Zhang and Maneesh Agrawala.
\newblock Adding conditional control to text-to-image diffusion models, 2023.

\bibitem{zhang2023controlvideo}
Yabo Zhang, Yuxiang Wei, Dongsheng Jiang, Xiaopeng Zhang, Wangmeng Zuo, and Qi Tian.
\newblock Controlvideo: Training-free controllable text-to-video generation.
\newblock {\em arXiv preprint arXiv:2305.13077}, 2023.

\end{thebibliography}
}

\end{document}


\title{Supplementary Materials for ``Ada-VE: Training-Free Consistent Video Editing Using Adaptive Motion Prior''}

\author{Tanvir Mahmud, Mustafa Munir, Radu Marculescu, and Diana Marculescu\\
The University of Texas at Austin
}
\maketitle

\appendix

\section{Limitations: Qualitative visualizations}
\begin{figure*}[t]
\centering
\includegraphics[width= 0.6\textwidth]{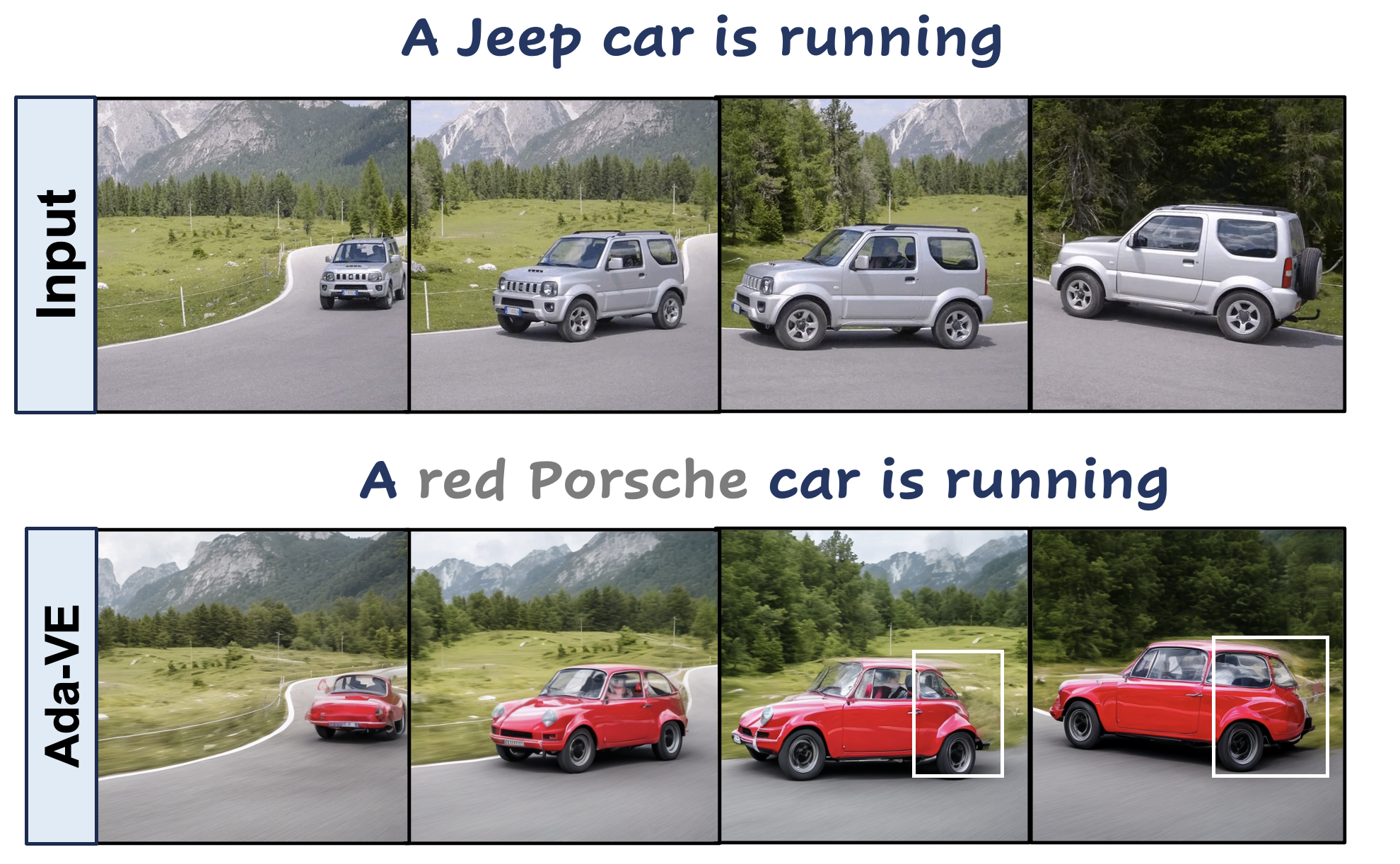}
\caption{\textbf{Visualizing failure cases:} Since our method is built on top of Plug-and-Play~\cite{pnpDiffusion2023}} diffusion, it cannot inherently follow the modified prompts to change the structure of the subject. Nevertheless, our key contributions are easy to adapt in most existing video editing baselines.
\label{fig:7}
\end{figure*}

We note that our method primarily built on top of the PnP~\cite{pnpDiffusion2023} image editing framework. PnP leverages structural guidance from the given video through extensive feature injection, which often fails to significantly alter the shape of subjects. We present a sample visualization of such failure cases in Fig.~\ref{fig:7}. It can be seen that the error mostly occurs when attempting to modify a Jeep car into a Porsche car due to their significant shape differences. These issues are largely inherited from the baseline PnP image editing method. Despite this, Ada-VE maintains consistent character across all frames, demonstrating its robustness in extending image models for video editing applications. Additionally, Ada-VE can be easily adapted to any image editing method due to its simple and general architecture.

\begin{figure*}[t]
\centering
\includegraphics[width= 0.6\textwidth]{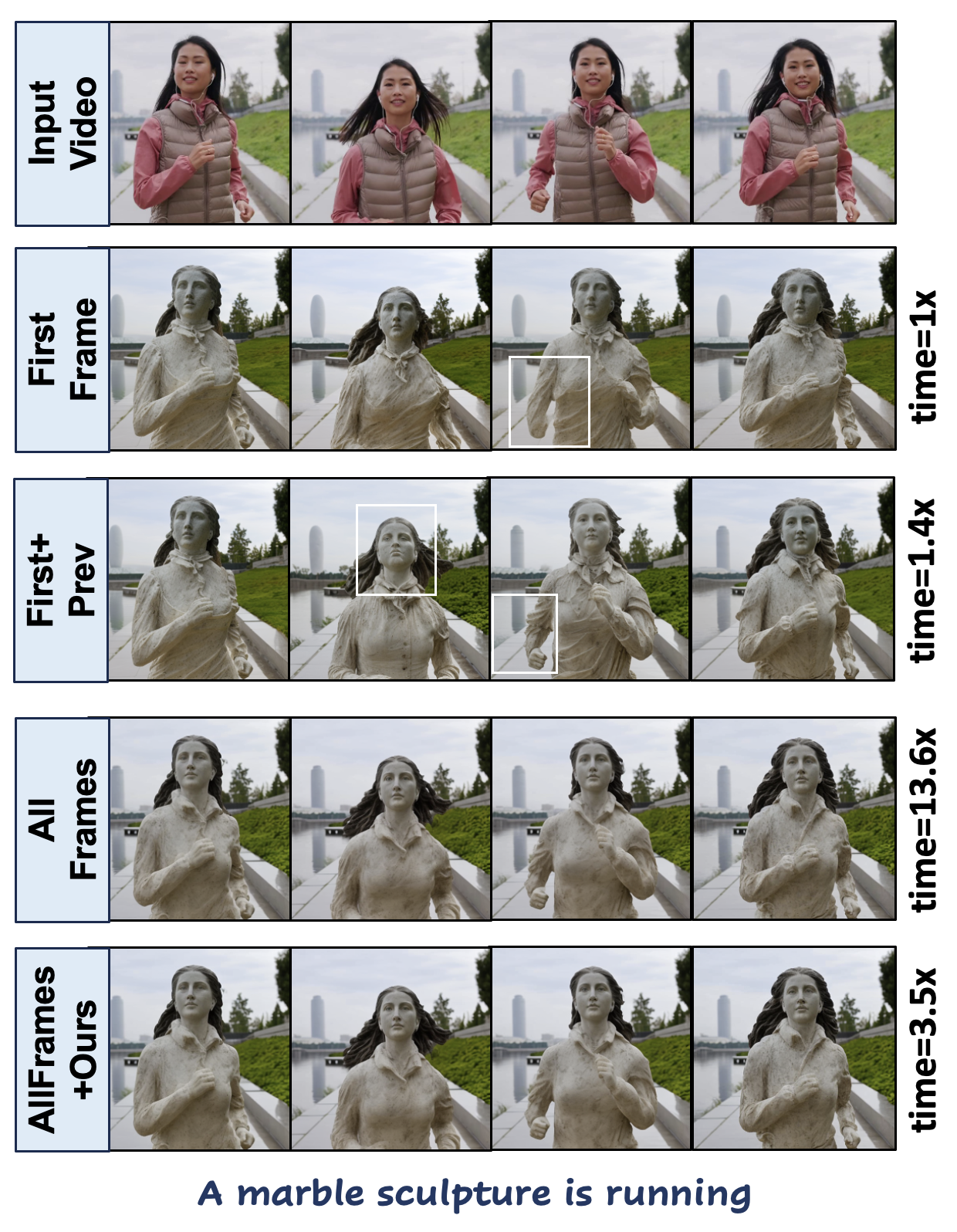}
\caption{Qualitative visualizations of various self-attention extensions: Increasing the number of frames in the extended self-attention significantly improves performance but comes with an extensive computational burden. Ada-VE significantly reduces the operational latency of full extensions while preserving visual quality by leveraging the motion prior of the guidance video. This allows for a substantial increase in the number of frames in joint editing, enabling efficient operation on longer duration videos. For this study, joint editing of a total of 40 frames was used, and we present a portion of these.}
\label{fig:8}
\end{figure*}

\section{Various extensions of Self-Attention: Qualitative visualizations}
We present sample qualitative visualizations of various self-attention extension mechanisms in Fig.~\ref{fig:8}. We observe consistent character generation when using KVs from a fixed set of frames across the videos. For instance, even when using only the first frame's KVs, we observe consistent character generation, but the visual quality deteriorates significantly, resulting in structural deformations. Integrating KVs from the immediate previous frames along with the first frame shows some improvement in structural deformations. However, using different KVs across the video leads to significant flickering and inconsistent character generation. For example, the woman's face is altered, and the background grass is no longer visible.

Using fully extended self-attention significantly improves visual quality and character consistency but increases latency by approximately 13 times due to the computational burden. We observe repeated features across frames, such as the background scenarios of the running woman, which consume a large portion of redundant computation. By integrating Ada-VE, we leverage the motion prior of the guidance video to drastically reduce computational overhead. Ada-VE achieves around a 4$\times$ speed-up for joint editing across these frames. Therefore, Ada-VE can potentially integrate a significantly larger number of frames in joint editing while using similar computational resources.

\section{Extensive Qualitative Visualizations}
The supplementary HTML page includes extensive qualitative visualizations on challenging examples, comparing Ada-VE with six state-of-the-art baseline methods. We also present results on ablation studies highlighting different self-attention extension mechanisms.

{\small
\bibliographystyle{ieee_fullname}
\bibliography{reference}
}